\PassOptionsToPackage{unicode}{hyperref}
\PassOptionsToPackage{hyphens}{url}
\PassOptionsToPackage{dvipsnames,svgnames,x11names}{xcolor}
\documentclass[
]{article}

\usepackage{amsmath,amssymb}
\usepackage{iftex}
\ifPDFTeX
  \usepackage[T1]{fontenc}
  \usepackage[utf8]{inputenc}
  \usepackage{textcomp} 
\else 
  \usepackage{unicode-math}
  \defaultfontfeatures{Scale=MatchLowercase}
  \defaultfontfeatures[\rmfamily]{Ligatures=TeX,Scale=1}
\fi
\usepackage{lmodern}
\ifPDFTeX\else  
\fi
\IfFileExists{upquote.sty}{\usepackage{upquote}}{}
\IfFileExists{microtype.sty}{
  \usepackage[]{microtype}
  \UseMicrotypeSet[protrusion]{basicmath} 
}{}
\makeatletter
\@ifundefined{KOMAClassName}{
  \IfFileExists{parskip.sty}{%
    \usepackage{parskip}
  }{
    \setlength{\parindent}{0pt}
    \setlength{\parskip}{6pt plus 2pt minus 1pt}}
}{
  \KOMAoptions{parskip=half}}
\makeatother
\usepackage{xcolor}
\ifx\paragraph\undefined\else
  \let\oldparagraph\paragraph
  \renewcommand{\paragraph}[1]{\oldparagraph{#1}\mbox{}}
\fi
\ifx\subparagraph\undefined\else
  \let\oldsubparagraph\subparagraph
  \renewcommand{\subparagraph}[1]{\oldsubparagraph{#1}\mbox{}}
\fi

\providecommand{\tightlist}{%
  \setlength{\itemsep}{0pt}\setlength{\parskip}{0pt}}\usepackage{longtable,booktabs,array}
\usepackage{calc} 
\usepackage{etoolbox}
\makeatletter
\patchcmd\longtable{\par}{\if@noskipsec\mbox{}\fi\par}{}{}
\makeatother
\IfFileExists{footnotehyper.sty}{\usepackage{footnotehyper}}{\usepackage{footnote}}
\usepackage{graphicx}
\makeatletter
\def\maxwidth{\ifdim\Gin@nat@width>\linewidth\linewidth\else\Gin@nat@width\fi}
\def\maxheight{\ifdim\Gin@nat@height>\textheight\textheight\else\Gin@nat@height\fi}
\makeatother
\setkeys{Gin}{width=\maxwidth,height=\maxheight,keepaspectratio}
\makeatletter
\def\fps@figure{htbp}
\makeatother
\newlength{\cslhangindent}
\newlength{\csllabelwidth}
\newlength{\cslentryspacingunit} 
\newenvironment{CSLReferences}[2] 
 {
  \setlength{\parindent}{0pt}
  \ifodd #1
  \let\oldpar\par
  \def\par{\hangindent=\cslhangindent\oldpar}
  \fi
  \setlength{\parskip}{#2\cslentryspacingunit}
 }%
 {}
\usepackage{calc}

\usepackage{algorithm}
\usepackage{algorithmic}
\usepackage{arxiv}
\usepackage{orcidlink}
\usepackage{amsmath}
\usepackage[T1]{fontenc}
\makeatletter
\@ifpackageloaded{caption}{}{\usepackage{caption}}
\AtBeginDocument{%
\ifdefined\contentsname
  \renewcommand*\contentsname{Table of contents}
\else
  \newcommand\contentsname{Table of contents}
\fi
\ifdefined\listfigurename
  \renewcommand*\listfigurename{List of Figures}
\else
  \newcommand\listfigurename{List of Figures}
\fi
\ifdefined\listtablename
  \renewcommand*\listtablename{List of Tables}
\else
  \newcommand\listtablename{List of Tables}
\fi
\ifdefined\figurename
  \renewcommand*\figurename{Figure}
\else
  \newcommand\figurename{Figure}
\fi
\ifdefined\tablename
  \renewcommand*\tablename{Table}
\else
  \newcommand\tablename{Table}
\fi
}
\@ifpackageloaded{float}{}{\usepackage{float}}
\floatstyle{ruled}
\@ifundefined{c@chapter}{\newfloat{codelisting}{h}{lop}}{\newfloat{codelisting}{h}{lop}[chapter]}
\floatname{codelisting}{Listing}

\makeatother
\makeatletter
\@ifpackageloaded{caption}{}{\usepackage{caption}}
\@ifpackageloaded{subcaption}{}{\usepackage{subcaption}}
\makeatother
\makeatletter
\@ifpackageloaded{tcolorbox}{}{\usepackage[skins,breakable]{tcolorbox}}
\makeatother
\makeatletter
\@ifundefined{shadecolor}{\definecolor{shadecolor}{rgb}{.97, .97, .97}}
\makeatother
\ifLuaTeX
  \usepackage{selnolig}  
\fi
\IfFileExists{bookmark.sty}{\usepackage{bookmark}}{\usepackage{hyperref}}
\IfFileExists{xurl.sty}{\usepackage{xurl}}{} 
\hypersetup{
  pdftitle={Online GentleAdaBoost - Technical Report},
  pdfauthor={Chapman Siu},
  pdfkeywords={online boosting, adaboost, gentleadaboost, machine
learning, classification},
  colorlinks=true,
  linkcolor={blue},
  filecolor={Maroon},
  citecolor={Blue},
  urlcolor={Blue},
  pdfcreator={LaTeX via pandoc}}

\newcommand{\runninghead}{A Preprint }
\title{Online GentleAdaBoost - Technical Report}
\author{
\textbf{Chapman Siu}\\Faculty of Engineering and Information
Technology\\University of Technology
Sydney\\\\\href{mailto:chapman.siu@student.uts.edu.au}{chapman.siu@student.uts.edu.au}}
\date{August 2023}
\begin{document}
\maketitle
\begin{abstract}
We study the online variant of GentleAdaboost, where we combine a weak
learner to a strong learner in an online fashion. We provide an approach
to extend the batch approach to an online approach with theoretical
justifications through application of line search. Finally we compare
our online boosting approach with other online approaches across a
variety of benchmark datasets.
\end{abstract}
{\bfseries \emph Keywords}
\def\sep{\textbullet\ }
online boosting \sep adaboost \sep gentleadaboost \sep machine
learning \sep 
classification

\ifdefined\Shaded\renewenvironment{Shaded}{\begin{tcolorbox}[breakable, frame hidden, borderline west={3pt}{0pt}{shadecolor}, sharp corners, interior hidden, enhanced, boxrule=0pt]}{\end{tcolorbox}}\fi

\hypertarget{sec-intro}{%
\section{Introduction}\label{sec-intro}}

Boosting algorithms belong to a class of ensemble classification
approaches which use weak assumptions on the learner to efficient manner
to improve performance. GentleBoost is an algorithm which was first
introduced as an alternative Adaboost approach which uses Newton steps
rather than exact optimization on each step (see Friedman, Hastie, and
Tibshirani 2000, p353). Unlike other AdaBoost variants, GentleBoost has
not received as much attention as it yields empirically inferior
performance compared with other Adaboost algorithms when used on a wide
range of benchmark datasets.

In machine learning, the ability to extend algorithms from a batch
setting to an online setting is an important topic. Online approaches
can operate on streams and use datasets which are too large to fit in
memory. In this technical report we provide an approach to extend
GentleBoost to the online setting through using line search. In addition
we perform experiments to demonstrate that the algorithm is
theoretically sound and has practical usecases.

\hypertarget{online-gentleboost}{%
\section{Online Gentleboost}\label{online-gentleboost}}

To describe the Online Gentleboost algorithm, we first describe the
Gentleboost algorithm for the two-class classification scenario. The
fitting procedure uses training data \((x_1, y_1), \dots, (x_n, y_n)\)
where \(x_i\) is a training instance vector and \(y_i \in \{-1, 1\}\).
Then define \(F(x) = \sum_1^M f_m(x)\) where every \(f_m(x)\) is some
weak classifier. Then the corresponding prediction is provided by
\(\text{sign}(F(x))\). For Gentleboost, it uses the \emph{exponential
criterion}, \(J(F) = E(\exp^{-yF(x)})\) for estimation of \(F(x)\).

Then if we use Newton steps for minimizing \(J(F)\)

\[\frac{\partial J(F(x) + f(x))}{\partial f(x)} \vert_{f(x)=0} = - E(\exp^{-yF(x)} y | x)\]

\[\frac{\partial^2 J(F(x) + f(x))}{\partial f(x)^2} \vert_{f(x)=0} =  E(\exp^{-yF(x)} | x)\text{ , since } y^2=1\]

The corresponding Newton update is

\[F(x) \leftarrow F(x) + \frac{E(\exp^{-yF(x)} y | x)}{E(\exp^{-yF(x)} | x)}\]

The GentleBoost algorithm is then summarised in
Table~\ref{tbl-gentleboost} shown below

\hypertarget{tbl-gentleboost}{}
\begin{longtable}[]{@{}
  >{\raggedright\arraybackslash}p{(\columnwidth - 0\tabcolsep) * \real{1.0092}}@{}}
\caption{\label{tbl-gentleboost}GentleBoost algorithm which is a
modified version of AdaBoost that uses Newton stepping rather than exact
optimization at each step}\tabularnewline
\toprule\noalign{}
\begin{minipage}[b]{\linewidth}\raggedright
\textbf{GentleBoost} (see Friedman, Hastie, and Tibshirani 2000, p353)
\textbar{}
\end{minipage} \\
\midrule\noalign{}
\endfirsthead
\toprule\noalign{}
\begin{minipage}[b]{\linewidth}\raggedright
\textbf{GentleBoost} (see Friedman, Hastie, and Tibshirani 2000, p353)
\textbar{}
\end{minipage} \\
\midrule\noalign{}
\endhead
\bottomrule\noalign{}
\endlastfoot
\begin{minipage}[t]{\linewidth}\raggedright
\begin{enumerate}
\def\labelenumi{\arabic{enumi}.}
\tightlist
\item
  Start with weights $w_i = 1/N, i = 1,2, \dots, N, F(x) =0$
\item
  Repeat for \(m = 1,2, \dots, M\):

  \begin{enumerate}
  \def\labelenumii{\alph{enumii}.}
  \tightlist
  \item
    Fit the regression function \(f_m(x)\) by weighted least-squares of
    \(y_i\) to \(x_i\) with weights \(w_i\).
  \item
    Update \(F(x) \leftarrow F(x) + f_m(x)\).
  \item
    Update \(w_i \leftarrow w_i \exp(-y_i f_m(x_i))\) and renormalize
  \end{enumerate}
\item
  Output the classifier
  \(\text{sign}(F(x)) = \text{sign}(\sum_{m=1}^M f_m(x))\).
\end{enumerate}
\end{minipage} \\
\end{longtable}

In our online boosting framework, the instances \((x_i, y_i)\) only
become available one at a time and the boosting algorithm must operate
in an online fashion as well. As such it is not possible for the
algorithm to determine the precise Newton Step at every instance.
Instead, we perform line search over Newton steps, which is known to
converge to the optimal Newton Step solution with sufficient small step.
The choice of the step size becomes a hyperparameter related to the
model, and removes the need to renormalize. The step size is chosen
based on the observation it needs to be proportional to
\(\exp(-y_i f_m(x_i))\) and bounded by the range of
\(\exp(-y_i f_m(x_i))\) to meet the Lipschwitz condition (Armijo 1966).
Since \(-1 \leq -y_i f_m(x_i) \leq 1\) then with the choice of
hyperparamter \(\alpha \in (0, \exp(1)-1)\) the step size
\(\hat{\alpha}\) is constructed as

\[  \hat{\alpha} = \begin{cases}
      \frac{1}{1+\alpha}, & \text{if}\ \text{sign}(-y_i f_m(x_i)) > 0 \\
      1+\alpha, & \text{otherwise}
    \end{cases}\]

As this approach uses a line search, any update function which
directionally moves the weight in the correct direction will be
suitable. The modified Online GentleBoost algorithm is summarised in
Table~\ref{tbl-online-gentleboost} shown below

\hypertarget{tbl-online-gentleboost}{}
\begin{longtable}[]{@{}
  >{\raggedright\arraybackslash}p{(\columnwidth - 0\tabcolsep) * \real{1.0092}}@{}}
\caption{\label{tbl-online-gentleboost}Online GentleBoost algorithm
which is a modified version of GentleBoost to allow for online
learning}\tabularnewline
\toprule\noalign{}
\begin{minipage}[b]{\linewidth}\raggedright
\textbf{Online GentleBoost}
\end{minipage} \\
\midrule\noalign{}
\endfirsthead
\toprule\noalign{}
\begin{minipage}[b]{\linewidth}\raggedright
\textbf{Online GentleBoost}
\end{minipage} \\
\midrule\noalign{}
\endhead
\bottomrule\noalign{}
\endlastfoot
\begin{minipage}[t]{\linewidth}\raggedright
\begin{enumerate}
\def\labelenumi{\arabic{enumi}.}
\tightlist
\item
  Start $F(x) =0$, with hyperparamter \(\alpha \in (0, \exp(1))\)
\item
  For incoming instance \(x_i, y_i\), reset weight \(w_i = 1\):
\item
  Repeat for \(m = 1,2, \dots, M\):

  \begin{enumerate}
  \def\labelenumii{\alph{enumii}.}
  \tightlist
  \item
    Fit the regression function \(f_m(x)\) by weighted least-squares of
    \(y_i\) to \(x_i\) with weights \(w_i\).
  \item
    Update \(F(x) \leftarrow F(x) + f_m(x)\).
  \item
    Update \(w_i \leftarrow \hat{\alpha} w_i\) and renormalize
    \textbar{}
  \end{enumerate}
\item
  Go back to 2. if there are additional instances
\item
  Finally output the classifier
  \(\text{sign}(F(x)) = \text{sign}(\sum_{m=1}^M f_m(x))\).
\end{enumerate}
\end{minipage} \\
\end{longtable}

\hypertarget{results}{%
\section{Results}\label{results}}

We use the benchmark datasets and approaches in the River (Montiel et
al. 2021) library to demonstrate the efficacy of our approach.

The model configuration uses the default settings and Hoeffding Trees
Hulten, Spencer, and Domingos (2001) as the ensemble approach for
AdaBoost (Oza and Russell 2001), Bagging (Oza and Russell 2001),
GentleBoost algorithms. We also compare our approach with ADWIN Bagging
Oza and Russell (2001), ALMA (Gentile 2000), Adaptive Random Forest
(Gomes et al. 2017), Aggregated Mondrian Forest (Mourtada, Gaiffas, and
Scornet 2019), Naive Bayes and Logistic Regression.

\hypertarget{tbl-results}{}
\begin{longtable}[]{@{}
  >{\raggedright\arraybackslash}p{(\columnwidth - 8\tabcolsep) * \real{0.3944}}
  >{\raggedright\arraybackslash}p{(\columnwidth - 8\tabcolsep) * \real{0.1549}}
  >{\raggedright\arraybackslash}p{(\columnwidth - 8\tabcolsep) * \real{0.1408}}
  >{\raggedright\arraybackslash}p{(\columnwidth - 8\tabcolsep) * \real{0.1690}}
  >{\raggedright\arraybackslash}p{(\columnwidth - 8\tabcolsep) * \real{0.1408}}@{}}
\caption{\label{tbl-results}Performance of Online GentleBoost compared
with other algorithms in River}\tabularnewline
\toprule\noalign{}
\begin{minipage}[b]{\linewidth}\raggedright
\end{minipage} & \begin{minipage}[b]{\linewidth}\raggedright
Bananas
\end{minipage} & \begin{minipage}[b]{\linewidth}\raggedright
Elec2
\end{minipage} & \begin{minipage}[b]{\linewidth}\raggedright
Phishing
\end{minipage} & \begin{minipage}[b]{\linewidth}\raggedright
SMTP
\end{minipage} \\
\midrule\noalign{}
\endfirsthead
\toprule\noalign{}
\begin{minipage}[b]{\linewidth}\raggedright
\end{minipage} & \begin{minipage}[b]{\linewidth}\raggedright
Bananas
\end{minipage} & \begin{minipage}[b]{\linewidth}\raggedright
Elec2
\end{minipage} & \begin{minipage}[b]{\linewidth}\raggedright
Phishing
\end{minipage} & \begin{minipage}[b]{\linewidth}\raggedright
SMTP
\end{minipage} \\
\midrule\noalign{}
\endhead
\bottomrule\noalign{}
\endlastfoot
ADWIN Bagging & 0.625967 & 0.823773 & 0.893515 & 0.999685 \\
ALMA & 0.506415 & 0.906404 & 0.8264 & 0.764986 \\
AdaBoost & 0.677864 & 0.880581 & 0.878303 & 0.999443 \\
Adaptive Random Forest & 0.88696 & 0.876608 & 0.907926 & 0.999685 \\
Aggregated Mondrian Forest & 0.884318 & 0.854517 & 0.888711 &
0.999874 \\
Bagging & 0.634082 & 0.840436 & 0.893515 & 0.999685 \\
GentleBoost & 0.61672 & 0.807464 & 0.880705 & 0.999685 \\
Hoeffding Tree & 0.642197 & 0.795635 & 0.879904 & 0.999685 \\
Logistic regression & 0.543019 & 0.822166 & 0.888 & 0.999769 \\
Naive Bayes & 0.61521 & 0.728741 & 0.884708 & 0.993484 \\
\end{longtable}

From the results above, we observe that GentleBoost generally performs
worse across all datasets except for the Phishing dataset, however it
demonstrates measureable uplift compared with the base weak learner
(i.e.~Hoeffding Tree).

More empirical evidence is required to verify this claim, though we note
this inferior results is consistent with the batch GentleBoost empirical
results which have been previously reported (see Friedman, Hastie, and
Tibshirani 2000, p365).

\hypertarget{conclusion}{%
\section{Conclusion}\label{conclusion}}

We have introduced Online Gentleboost, an extension of the original
batch Gentleboost approach via line search. We have justified our
approach theoretically and demonstrated empirically that Gentleboost
does indeed improve upon the weak learner.

\hypertarget{references}{%
\section*{References}\label{references}}
\addcontentsline{toc}{section}{References}

\hypertarget{refs}{}
\begin{CSLReferences}{1}{0}
\leavevmode\vadjust pre{\hypertarget{ref-Armijo1966}{}}%
Armijo, Larry. 1966. {``{Minimization of functions having Lipschitz
continuous first partial derivatives.}''} \emph{Pacific Journal of
Mathematics} 16 (1): 1--3.

\leavevmode\vadjust pre{\hypertarget{ref-MOA2010}{}}%
Bifet, Albert, Geoff Holmes, Richard Kirkby, and Bernhard Pfahringer.
2010. {``MOA: Massive Online Analysis.''} \emph{J. Mach. Learn. Res.} 11
(August): 1601--4.

\leavevmode\vadjust pre{\hypertarget{ref-Friedman2000}{}}%
Friedman, Jerome, Trevor Hastie, and Robert Tibshirani. 2000. {``Special
Invited Paper. Additive Logistic Regression: A Statistical View of
Boosting.''} \emph{The Annals of Statistics} 28 (2): 337--74.
\url{http://www.jstor.org/stable/2674028}.

\leavevmode\vadjust pre{\hypertarget{ref-NIPS2000_d072677d}{}}%
Gentile, Claudio. 2000. {``A New Approximate Maximal Margin
Classification Algorithm.''} In \emph{Advances in Neural Information
Processing Systems}, edited by T. Leen, T. Dietterich, and V. Tresp.
Vol. 13. MIT Press.
\url{https://proceedings.neurips.cc/paper_files/paper/2000/file/d072677d210ac4c03ba046120f0802ec-Paper.pdf}.

\leavevmode\vadjust pre{\hypertarget{ref-gomes2017adaptive}{}}%
Gomes, Heitor M, Albert Bifet, Jesse Read, Jean Paul Barddal, Fabricio
Enembreck, Bernhard Pfharinger, Geoff Holmes, and Talel Abdessalem.
2017. {``Adaptive Random Forests for Evolving Data Stream
Classification.''} \emph{Machine Learning} 106: 1469--95.

\leavevmode\vadjust pre{\hypertarget{ref-hulten2001}{}}%
Hulten, Geoff, Laurie Spencer, and Pedro Domingos. 2001. {``Mining
Time-Changing Data Streams.''} In \emph{Proceedings of the Seventh ACM
SIGKDD International Conference on Knowledge Discovery and Data Mining},
97--106. KDD '01. New York, NY, USA: Association for Computing
Machinery. \url{https://doi.org/10.1145/502512.502529}.

\leavevmode\vadjust pre{\hypertarget{ref-montiel2021river}{}}%
Montiel, Jacob, Max Halford, Saulo Martiello Mastelini, Geoffrey
Bolmier, Raphael Sourty, Robin Vaysse, Adil Zouitine, et al. 2021.
{``River: Machine Learning for Streaming Data in Python.''}

\leavevmode\vadjust pre{\hypertarget{ref-mourtada2019amf}{}}%
Mourtada, Jaouad, Stephane Gaiffas, and Erwan Scornet. 2019. {``Amf:
Aggregated Mondrian Forests for Online Learning.''} \emph{arXiv Preprint
arXiv:1906.10529}.

\leavevmode\vadjust pre{\hypertarget{ref-oza01a}{}}%
Oza, Nikunj C., and Stuart J. Russell. 2001. {``Online Bagging and
Boosting.''} In \emph{Proceedings of the Eighth International Workshop
on Artificial Intelligence and Statistics}, edited by Thomas S.
Richardson and Tommi S. Jaakkola, R3:229--36. Proceedings of Machine
Learning Research. PMLR.
\url{https://proceedings.mlr.press/r3/oza01a.html}.

\end{CSLReferences}

\end{document}